\documentclass[a4paper, 12pt, twoside]{article}
\input{JADTstyle.cls}

\usepackage{subcaption}

\title{\fontsize{18}{21.6}\textbf{A KAN-based Non-linear Factorization for Embedding Layer Compression}}

\author{Oliver Giudice}
\author{Anna Russo Russo}

\affil{Applied Research Team, IT Dept. Banca d'Italia \footnote{Views and opinions are those of the authors and do not necessarily reflect the official policy or position of Banca d’Italia. Authors listed in alphabetical order.} $-$ oliver.giudice@bancaditalia.it, anna.russorusso@bancaditalia.it}

\date{\vspace*{-3em}}

\begin{document}

\maketitle

\fancyhead[CE]{\textsc{\footnotesize{Oliver Giudice, Anna Russo Russo}}}
\fancyhead[CO]{\textsc{\footnotesize{A KAN-based non-linear factorization for Embedding Layer Compression}}}

\selectlanguage{english} 

\setlength{\parindent}{\z@}
\setlength{\baselineskip}{11pt}


\vspace*{2pt}
\abstract*{Abstract}

\begin{footnotesize}{Compressing the embedding layer in Natural Language Processing (NLP) models is a critical yet challenging problem: embeddings often account for a substantial fraction of the model parameters while being highly sensitive to information loss. On the one hand effective compression can significantly reduce memory footprint and improve deployment efficiency, but on the other hand even small distortions in the embedding space may propagate through the entire neural network thus degrading downstream performance.

Among existing embedding compression approaches, low-rank factorization provides a simple and mathematically principled solution by approximating the embedding matrix with a reduced-rank decomposition. However, its practical effectiveness is often limited, as the resulting linear bottleneck constrains representational capacity. To overcome these limitations, we propose an enhanced factorization scheme that increases expressive power by introducing learnable non-linear projections within the compressed embedding layer. Our method is inspired by the Kolmogorov–Arnold Networks (KAN) paradigm, which replaces fixed linear transformations with learnable univariate functions. Despite promising results in other domains, applications of KANs in NLP remain scarce and have so far been largely limited to final or task-specific layers, rather than core representational components such as embedding layers.

We evaluate the proposed approach by integrating the compressed embedding layer into the Skip-gram framework and comparing it against a standard low-rank baseline under matched parameter budgets on the Word Analogy and Semantic Similarity benchmarks. We further introduce an effective initialization strategy from a pre-trained, uncompressed embedding layer. The behavior of the model is analyzed both immediately after initialization and after a brief post-initialization refinement phase, considering training with the standard Skip-gram objective as well as variants augmented with distillation from the original embedding layer.
Overall, our results show that augmenting low-rank factorization with learnable non-linear projections effectively mitigates the limitations of linear bottlenecks, yielding more faithful compressed embeddings under fixed dimensionality and suggesting a promising direction for embedding layer compression beyond purely linear methods.

}\end{footnotesize}

\vspace*{14pt}{\textbf{Keywords:}} \begin{footnotesize}{Natural Language Processing, Embedding Compression, Low-Rank Factorization, Kolmogorov-Arnold Networks}\end{footnotesize}


\setlength{\parskip}{6pt}
\setlength{\baselineskip}{14pt}

\section{Introduction}
Word embeddings have become an essential building block of modern Natural Language Processing (NLP), enabling the representation of words as dense vectors that capture semantic and syntactic relationships. However, as vocabulary sizes grow, the parameter count and memory footprint of standard embedding layers become a significant bottleneck, limiting scalability and hindering deployment in memory-constrained environments. To address these challenges, a broad family of compressed embedding techniques has been developed, among which low-rank factorization \citep{lan_2019, lioutas-2020-improving}, hashing \citep{chen_2015_compressing}, quantization \citep{shi_2018_structured, shu_2017_compressing}, distillation learning \citep{gao_2023_distilcse}, pruning or weight sharing strategies \citep{chen_2018_groupreduce, suzuki_2016_learning} and others \citep{balazy_2021_direction}.
Within this landscape, low-rank factorization provides a principled and widely used baseline, being broadly adopted in NLP and other machine learning settings beyond embedding compression. In the standard approach, the factorization is strictly linear, as it decomposes the embedding matrix into two smaller matrices.

In this work we propose a novel low-rank factorization technique based on Kolmogorov–Arnold Networks (KANs). KANs, recently popularized by \cite{liu_kan_2024}, are neural architectures that employ learnable, non-linear, univariate activation functions. Compared to traditional Multi-Layer Perceptrons (MLPs), this architecture appears to exhibit more favorable scaling-law constants and has been empirically shown to achieve higher accuracy and improved interpretability at a fixed parameter budget. The proposed factorization technique aims at overcoming the implicit linear bottleneck of the standard low-rank factorization by replacing the second linear projection with a KAN-based mapping in which each output dimension is computed as a sum of learned non-linear transformations of the individual input coordinates.

To the best of our knowledge, KAN-style learnable activation functions have not been explored as a mechanism for embedding compression. The low‑rank factorization setting provides a clean way to isolate their contribution.
To this end, we introduce a KAN-based factorized embedding layer along with two initialization schemes and we evaluate it within the Skipgram framework against a standard low-rank factorization.

The paper is structured as follows: Section \ref{sec:related_work} discusses related work on KANs, while Section \ref{sec:background_kan} provides preliminary background and notation; Section \ref{sec:main} explains our original KAN-based factorization technique. The experimental framework used for evaluation is described in Section \ref{sec:experiments}, while the experimental results are reported and discussed in Section \ref{sec:results}. Lastly, Section \ref{sec:conclusions} offers final considerations and future research directions for the employment of KANs in NLP.

\section{Related Work}
\label{sec:related_work}

Since their introduction, KANs have been explored across a wide range of domains, including time-series \citep{han_are_2025, vaca-rubio_kolmogorov-arnold_2024, xu_kolmogorov-arnold_2024}, computer vision \citep{cheon_demonstrating_2024, yu_residual_2025}, reinforcement learning \citep{guo_kan_2024, kich_kolmogorov-arnold_2024}, speech \citep{xu_effective_2025}, and physics-informed modeling \citep{howard_finite_2024, so_higher_2024, wang_kolmogorov_2025}, often as replacements for fully-connected blocks within standard architectures (e.g., CNNs \citep{bodner_convolutional_2025}, RNNs/LSTMs \citep{genet_tkan_2024}, GNNs \citep{xu_enhancing_2025, zhang_graphkan_2024}, Transformers \citep{yang_kolmogorov-arnold_2024, genet_temporal_2024}, and U-Nets \citep{li_u-kan_2024}).
A substantial body of follow-up work has focused on modifying the functional form of the learnable univariate components, which are implemented as spline-based parametrizations in the original formulation. Among the proposed replacement, we mention rational functions \citep{aghaei_rkan_2024}, radial basis functions \citep{li_kolmogorov-arnold_2024}, sinusoidal functions \citep{reinhardt_sinekan_2025}, Rectified Linear Units \citep{qiu_relu-kan_2024}.
Alternative aggregation functions in place of sums have also been tested \citep{liu_kan_2_2024, altarabichi_rethinking_2024}, as well as regularization and parameter-reduction strategies, such as pruning, dropout, parameter sharing, bottleneck designs, Mixture of Experts \citep{fan_shift_2025, altarabichi_dropkan_2024, drokin_kolmogorov-arnold_2024, yang_kolmogorov-arnold_2024}.
At the same time, recent studies \citep{yu_kan_2024, zeng_kan_2024, hou_kolmogorov-arnold_2025} have raised concerns about the fairness of empirical comparisons and the scope of KAN advantages, suggesting that performance gains may be confined to settings involving smooth, regular functions, small-to-medium model sizes, or the use of symbolic regression.

To the best of our knowledge, applications of KANs to the field of Natural Language Processing (NLP) are still scarce in literature; probably because the lack of smoothness in the involved inputs and the training instability of KANs make it hard to deploy them in state-of-the-art models. Among the existing applications, \citet{imran_fourier_2024} highlight the advantages of employing a KAN-based classification head on top of different language models for sentiment analysis, topic classification, question classification and language identification tasks.

\section{Background}
\label{sec:background_kan}

Kolmogorov-Arnold Networks (KANs) are neural architectures inspired by the Kolmogorov-Arnold representation theorem. The theorem states that any smooth function can be represented as a finite composition of continuous, but not necessarily smooth, univariate functions combined through summation.
Building on this idea, \cite{liu_kan_2024} proposed a neural network architecture in which each layer output is computed as a sum of learnable smooth univariate functions.

A KAN layer with index $l$ maps an input vector $x^{(l)} \in \mathbb{R}^{n_l}$ to an output vector $x^{(l+1)} \in \mathbb{R}^{n_{l+1}}$ by associating a learnable univariate function $\phi^{(l)}_{ij} : \mathbb{R} \to \mathbb{R}$ to each directed connection from input coordinate $i$ to output coordinate $j$, and aggregating contributions by summation:
\begin{equation}
x^{(l+1)}_j \;=\; \sum_{i=1}^{n_l} \phi^{(l)}_{ij}(x^{(l)}_i),
\qquad j = 1,\dots,n_{l+1}.
\label{eq:kan_layer}
\end{equation}
Equivalently, the layer can be viewed as an $n_l \times n_{l+1}$ matrix of univariate functions $\Phi^{(l)} = (\phi^{(l)}_{ij})$, applied coordinate-wise and summed across inputs.
Consequently, a KAN with $L$ layers is obtained through composition:
\begin{equation}
\mathrm{KAN}(x) \;=\; \left(\Phi^{(L-1)} \circ \Phi^{(L-2)} \circ \cdots \circ \Phi^{(0)}\right)(x),
\label{eq:kan_network}
\end{equation}
where each $\Phi^{(l)}$ denotes the non-linear mapping defined by Eq.~\eqref{eq:kan_layer}.
In the original formulation, each activation function $\phi^{(l)}_{ij}$ is parametrized as the sum of a base function and a spline component:
$\phi^{(l)}_{ij}(x_i^{(l)}) \;=\; \gamma^{(l)}_{ij} \, f(x_i^{(l)}) \;+\; \mathcal{S}^{(l)}_{ij}(x_i^{(l)})$,
where $f$ is a fixed basis function (e.g., a sigmoid-like function), $\gamma^{(l)}_{ij}$ is a learnable coefficient, and $\mathcal{S}^{(l)}_{ij}$ is a learnable spline of order $k$ defined on a grid of size $g$. However, the definition of KAN layers does not depend on a specific functional family and, as highlighted in Section \ref{sec:related_work}, several different parametrizations are possible.

\section{Methodology}
\label{sec:main}
\begin{figure}[!t]
\begin{subfigure}{\textwidth}
  \centering
  \includegraphics[width=.45\linewidth]{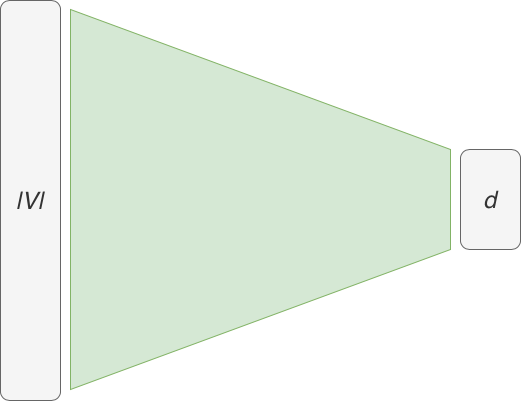}
  \caption{}
  \label{fig:sfig1}
\end{subfigure}\\[1ex]%
\begin{subfigure}{.5\textwidth}
  \centering
  \includegraphics[width=.9\linewidth]{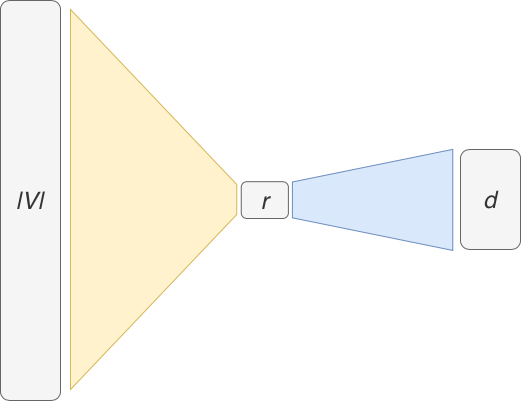}
  \caption{}
  \label{fig:sfig2}
\end{subfigure}
\begin{subfigure}{.5\textwidth}
  \centering
  \includegraphics[width=.9\linewidth]{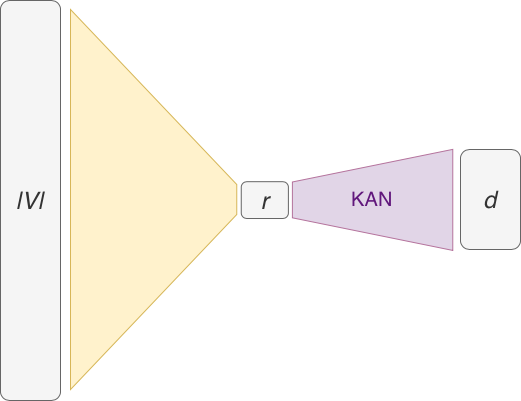}
  \caption{}
  \label{fig:sfig3}
\end{subfigure}
    \caption{Embeddings layers to embed vocabulary $V$ into a space of size $d$: (a) a standard full-rank embedding layer; (b) a low-rank factorized embedding layer of rank $r$; (c) a KAN-based factorized embedding layer of rank $r$.}
    \label{fig:factorized_embeddings_comparison}
\end{figure}

Given a vocabulary $V$ and embedding size $d$, a standard embedding matrix $E \in \mathbb{R}^{|V|\times d}$ can be approximated by a rank-$r$ factorization $E \approx A B$, where $A \in \mathbb{R}^{|V|\times r}$ and $B \in \mathbb{R}^{r \times d}$ with $r \ll d$.
In our method, we retain this factorized structure but replace the second linear mapping with the learnable non-linear transformation provided by a KAN layer of the same shape, namely with $r$ inputs and $d$ outputs:
$$Ex \approx KAN(Wx)$$
where $x$ denotes the one-hot vector of a vocabulary item and $W \in \mathbb{R}^{|V|\times r}$ plays the role of the first factor in the low-rank decomposition. 
Figure \ref{fig:factorized_embeddings_comparison} compares visually the proposed factorized KAN-based embedding layer against a standard embedding layer and a low-rank factorized embedding layer.

A practical advantage of low-rank factorized embeddings is that they admit a straightforward initialization. Given a pre-trained embedding matrix $E \in \mathbb{R}^{|V| \times d}$, the low-rank decomposition with minimum reconstruction error can be obtained through a singular-value decomposition (SVD), yielding matrices $A \in \mathbb{R}^{|V| \times r}$ and $B \in \mathbb{R}^{r \times d}$.
To enable an analogous initialization mechanism for our KAN-based factorized embeddings, we develop two strategies, both starting from the SVD of $E$. In each case, the matrix $A$ is used to initialize the linear transformation encoded in matrix $W$ that precedes the KAN layer. Since KAN activation functions operate within a fixed activation interval, defined by the splines grid, we first apply a scaling transformation controlled by a constant $\alpha>0$, deriving $A' = \alpha A$ and $B' = B/\alpha$. Then we initialize $W$ from $A'$.
The two methods differ in how the parameters of the KAN layer itself are initialized.

\paragraph{KAN-SVD initialization}
The first strategy aims to closely reproduce the linear behaviour of the SVD-based low‑rank factorization at initialization. After setting $W = A'$, each KAN activation function $\phi_{ij}$ is initialized so that, within the activation interval, it approximates a linear function with slope given by the corresponding entry of $B'$, i.e. $\phi_{ij}(y) \approx b'_{ij}\cdot y$. This can be achieved, for instance, by performing a least-squares regression over a small set of points within the activation interval.
We refer to this initialization as \emph{KAN-SVD}.

\paragraph{KAN-BCD initialization}
The second strategy adopts a block coordinate descent (BCD) scheme that alternates between updating the parameters of the KAN layer and the linear layer $W$, minimizing the reconstruction loss over the vocabulary $V$: $$\mathcal{L} = \sum_{x \in V} \|E\cdot x - KAN(W \cdot x)\|^2$$
At each iteration, we freeze $W$ and update the KAN parameters by solving a ridge-regularized least-squares problem. This problem admits a closed-form solution because the spline parameters enter linearly in the output of the KAN layer.
Then, we freeze the KAN layer and optimize $W$ to minimize $\mathcal{L}$ via stochastic gradient descent over a fixed number of steps.
The process continues until the relative improvement in $\mathcal{L}$ between consecutive iterations falls below a tolerance threshold.
Note that this strategy does not rely on the matrix $B$, but explores a more flexible non-linear reconstruction regime.
We denote this method as \emph{KAN‑BCD}.

\section{Experimental setup}
\label{sec:experiments}
We compare our KAN-based factorized embeddings against low-rank factorized embeddings of the same rank within the task of learning Skipgram-like word embeddings. The pre-trained embedding vectors to be compressed are derived by training a standard full-rank Skipgram model.
We assess information loss after compression using two intrinsic evaluation tasks: word analogies and semantic similarity. Both tasks are evaluated (i) immediately after compression, and (ii) after a short post-compression adaptation phase, in which the factorized model is fine-tuned on the same data used to train the full-rank embeddings. Following \citet{lioutas-2020-improving}, we compare two adaptation strategies: training with a standard Skipgram loss or including a feature distillation term from the original full-rank embeddings.

Throughout the rest of the paper, we denote low-rank factorization as \emph{LF}, and the two KAN-based factorized embeddings as \emph{KF-SVD} and \emph{KF-BCD}, corresponding respectively to the KAN-SVD and KAN-BCD initialization strategies. When discussing the effect of the post-compression adaptation phase, we use the suffix \emph{+} to distinguish the distilled models.

\subsection{Training data}
We train a standard Skipgram model with negative sampling following the original formulation of \citet{mikolov_distributed_2013}: this is a binary classifier trained to recognize whether two tokens are likely to be in the same context or not.
We use a corpus consisting of a 120M words subset of Wikipedia. After removing punctuation and stop words and discarding tokens with fewer than 100 occurrences, we obtain a vocabulary of approximately 37k tokens.
The training dataset is generated by scanning the corpus and collecting up to four target–context pairs per occurrence, using a context window of size 2.
Each target–context pair observed within the context window constitutes a positive training example, while negative examples are obtained by pairing the same target token with tokens randomly sampled from the vocabulary, through the frequency‑based sampling distribution recommended in \citet{mikolov_distributed_2013}. For each positive target-context pair, three negative examples are generated.

\subsection{Hyperparameters and implementation details}
We compress embeddings of size $d = 50$ with factorized embeddings of rank $r = 25$. As commonly done in Skipgram models, the same embedding layer is shared between target and context representations.

For the implementation of the KAN-based factorized embedding, we use the \texttt{fastkan} library, which employs Radial Basis Functions (RBFs) in place of splines. This choice is supported by \citet{li_kolmogorov-arnold_2024}, who shows that RBFs can approximate cubic B-splines, while being substantially more efficient. In this implementation, a KAN layer takes the form:
\begin{equation}
\label{eq:rbf}
    \Phi(x)  = \sum_{m=1}^{g} C^m \psi(\|x - p_m\|)
\end{equation}
where $\psi$ is a Gaussian kernel, $p_m$ are fixed grid points, and $C^m \in \mathbb{R}^{r\times d}$ are learnable parameters.
We define a grid of $g=10$ points spaced uniformly in the activation interval $[-1, 1]$.
The number of learnable parameters in each model is reported in Table \ref{tab:n_parameters_tested_models}.

\begin{table}
    \caption{Number of learnable parameters and associated compression rate with respect to the full-rank embeddings for the tested models.}
    \centering
    \begin{tabular}{c|r|r}
        Models & N. parameters & Compression rate \\
        \hline
        \textit{LF} & 935,150 & 2.00x \\
        \textit{KF-SVD, KF-BCD} & 946,400 & 1.97x\\
    \end{tabular}
    \label{tab:n_parameters_tested_models}
\end{table}

\paragraph{Initialization}
The matrices $A, B$ obtained via SVD of the full-rank embedding matrix $E$, are scaled by a constant factor $\alpha$ set to the inverse of the 99-percentile of $\vert A \vert$.

For the KAN-SVD strategy, the KAN layer is initialized to emulate a linear transformation with weight matrix $B'$. The RBF-based implementation that we use allows to do that by interpolating the identity function at grid points $p_1, \ldots, p_g$. Following the definition given in Eq. \eqref{eq:rbf}, the interpolating weights $c^*$ can be computed by solving the linear system $Qc^* = v$, where $Q_{hk} = \psi(\|p_h - p_k\|)$ and $v_k = p_k$ for $h, k = 1 \ldots g$. Such defined $Q$ is non-singular \citep{micchelli_1986}, thus the system is guaranteed to admit a solution. We then set $C_m = c^*_m B'$ for all $m=1, \ldots, g$.

In the context of the KAN-BCD initialization, the KAN parameters are optimized with a ridge coefficient equal to $10^{-4}$, while the linear layer is updated through the \emph{Adam} optimizer with learning rate $10^{-3}$ and for $20$ steps. We follow the BCD scheme up to $200$ iterations or until the relative improvement in reconstruction loss falls below $10^{-4}$.

\paragraph{Post-compression adaptation}
We fine-tune the factorized models using the Adam optimizer with learning rate $10^{-4}$.
When feature distillation is enabled, we add a penalty based on the cosine similarity between the current embedding $w_t$ and the original embedding $w'_t$: $\mathcal{L}_{FD}(t) = 1 - \frac{w_t \cdot w'_t}{\lVert w_t\rVert \lVert w'_t\rVert}$, for token $t$.
The loss for a target-context pair $(t, c)$ becomes:
 \begin{equation*}
     (1 - \lambda)   \mathcal{L}_{SG}(t, c) + \frac{\lambda}{2} (\mathcal{L}_{FD}(t) + \mathcal{L}_{FD}(c))
 \end{equation*}
where $\mathcal{L}_{SG}$ is the regular Skipgram loss function and $\lambda = 0.3$, following \citep{lioutas-2020-improving}.

\subsection{Embedding evaluation}
We evaluate analogy performance on the \emph{Google Analogy} dataset \citep{mikolov_efficient_2013}, containing 19 544 questions across semantic and syntactic relations, using the standard \emph{3CosAdd} formulation. Semantic similarity is assessed on \emph{WordSim-353} \citep{wordsim} and \emph{SimLex-999} \citep{simlex}, using Spearman correlation.

\subsection{Hardware and software environment}
All experiments were conducted on a single NVIDIA V100 GPU (16 GB) on the Ubuntu Linux operating system. Models were implemented in Python 3.10 with PyTorch 2.3.0, Hugging Face Transformers 4.45.2 and CUDA 12.4.

\section{Results}
\label{sec:results}
Table~\ref{tab:results_0} summarizes the performance of the evaluated models immediately after compression. As expected, low‑rank factorization (\emph{LF}) entails a substantial loss of information: its accuracy on word analogy tasks drops by almost $50\%$ relative to the full-rank baseline, and the overall average reduction is $28.8\%$.
\emph{KF-SVD} performs even worse ($31.2\%$ average reduction), which is consistent with its design: since it is initialized to closely replicate the behaviour of the compressed \emph{LF} embedding, it inherits and potentially amplifies its errors.
In contrast, \emph{KF-BCD} embeddings retain a considerably larger fraction of semantic structure. Their average performance reduction is only $16.4\%$, nearly halving the degradation observed for \emph{LF}. On analogy tasks and WordSim‑353, the same pattern emerges: \emph{LF} suffers a $39\%$ reduction on average, whereas \emph{KF-BCD} exhibits only $19\%$.
A different trend is observed on SimLex‑999. Both \emph{LF} and \emph{KF-SVD} achieve slightly higher correlation scores than the full-rank model. This suggests that SimLex may emphasize semantic relations that are underrepresented in the corpus used to train the original full‑rank Skipgram embeddings. Notably, this behaviour persists after adaptation: post‑compression fine‑tuning does not increase correlation on SimLex (Table~\ref{tab:results_3}), supporting the hypothesis that the discrepancy stems from corpus–dataset mismatch rather than from compression effects.

\begin{table}[!htbp]
    \caption{Scores achieved immediately after compression on the evaluation datasets. Columns \textit{syn. analogies accuracy} and \textit{sem. analogies accuracy} report the accuracy scored on the Google Analogy datasets, respectively for syntactical and semantical relations; columns \textit{wordsim correlation} and \textit{simlex correlation} report the Spearman correlation scored respectively on the WordSim-353 and SimLex-999 datasets. The last column displays the average relative reduction in performance with respect to the full-rank model.}
    \centering
        \begin{tabular}{l|r|r|r|r|r}
             & syn. analogies  & sem. analogies & wordsim & simlex & avg. performance \\
            Model & accuracy & accuracy & correlation & correlation & reduction (\%) \\
            \hline
            full-rank & 0.203 & 0.224 & 0.671 & 0.225 & - \\
            \hline
            LF & 0.100 & 0.103 & 0.585 & \textbf{0.230} & 28.8 \\
            KF-SVD & 0.093 & 0.091 & 0.585 & 0.229 & 31.2 \\
            KF-BCD & \textbf{0.149} & \textbf{0.161} & \textbf{0.652} & 0.207 & \textbf{16.4}\\
        \end{tabular}
    \label{tab:results_0}
\end{table}

Table~\ref{tab:results_3} reports the best performance achieved within three epochs of post-compression adaptation. Overall, the adaptation phase reduces the average gap to the full-rank baseline across all models, although the gains are not uniform. In the end, all KAN-based variants perform better on average than the standard low-rank factorization, with \emph{KF-SVD+} attaining the lowest overall average reduction (11.7\%) despite starting from the weakest post-compression initialization. These results suggest that the additional functional expressivity introduced by the KAN layer provides a more effective capacity for recovering task-relevant structure during fine-tuning.
In contrast, \emph{KF-BCD} and \emph{KF-BCD+} show only moderate improvements: this may reflect a less favourable optimization landscape induced by the BCD initialization.
The distillation loss is generally beneficial for both KAN-based variants, whereas it consistently harms adaptation for \emph{LF} embeddings.

\begin{table}[!htbp]
    \caption{Best scores achieved over 3 epochs of post-compression adaptation. Columns as in Table \ref{tab:results_0}.}
    \centering
        \begin{tabular}{l|r|r|r|r|r}
             & syn. analogies  & sem. analogies & wordsim & simlex & avg. perf. \\
            Model & accuracy & accuracy & correlation & correlation & reduction (\%) \\
            \hline
            full-rank & 0.203 & 0.224 & 0.671 & 0.225 & - \\
            \hline
            LF & 0.155 & 0.134 & 0.625 & \textbf{0.230} & 17.1 \\
            LF+ & 0.111 & 0.103 & 0.590 & \textbf{0.230} & 27.3 \\
            KF-SVD & \textbf{0.163} & 0.128 & 0.629 & 0.229 & 16.8 \\
            KF-SVD+ & 0.152 & \textbf{0.174} & \textbf{0.663} & 0.229 & \textbf{11.7} \\
            KF-BCD & 0.149 & 0.159 & 0.653 & 0.214 & 15.8 \\
            KF-BCD+ & 0.149 & 0.166 & 0.652 & 0.207 & 15.8 \\
        \end{tabular}
    \label{tab:results_3}
\end{table}

\section{Conclusion and future work}
\label{sec:conclusions}

Our results indicate that KAN‑based factorized embeddings constitute a viable and competitive alternative to traditional low-rank factorizations, as learnable non‑linearities introduced by KAN layers enable to recover a substantial portion of the expressiveness lost through compression.

The two initialization schemes we proposed exhibit complementary advantages. KAN‑SVD excels at recovering information directly from the full‑rank embedding matrix, while KAN‑BCD offers a more effective starting point for post-compression adaptation or, possibly, downstream fine‑tuning. At the same time, the heterogeneous behaviour observed across the two post‑compression adaptation strategies highlights that training stability remains a central challenge for KAN‑based embedding layers. These findings point to the need for a more systematic understanding of how initialization, optimization hyperparameters, and fine‑tuning procedures interact with KAN architectures.

Future work will extend this study along two directions. First, evaluating KAN‑based embeddings on downstream NLP tasks will clarify whether the advantages observed in intrinsic evaluation translate into gains in practical applications. Second, integrating KAN‑based factorized embeddings into larger architectures, most notably Transformer models, will shed light on their compatibility with modern NLP pipelines.

Overall, our study suggests that KAN-based non-linear factorization is a promising direction for making embedding compression more expressive without sacrificing parameter efficiency.

\bibliography{main}         

\end{document}